\documentclass[letterpaper, 10 pt, journal, twoside]{IEEEtran}
\usepackage{amsmath}
\usepackage{amsfonts}
\usepackage{algorithmic}
\usepackage{algorithm}
\usepackage{array}
\usepackage[caption=false,font=normalsize,labelfont=sf,textfont=sf]{subfig}
\usepackage{textcomp}
\usepackage{stfloats}
\usepackage{url}
\usepackage{verbatim}
\usepackage{graphicx}
\usepackage{cite}

\usepackage{bm}

\usepackage{enumitem}

\usepackage{glossaries}

\usepackage{xcolor}

\usepackage{hyperref}
\hypersetup
{
	pdftitle = {Landing Controller for Quadruped Robots},
	pdfauthor = {Francesco Roscia, Andrea Del Prete, Claudio Semini, Michele Focchi},
	pdfsubject = {RA-L manuscript},
	pdfkeywords = {legged robots, aerial locomotion, and safe landing},
	pdftoolbar = true,
}

\usepackage{amsthm}

\usepackage{tensor}

\usepackage{amssymb}

\usepackage{graphicx}

\usepackage[export]{adjustbox}

\newacronym{com}{CoM}{Center of Mass}
\newacronym{cop}{CoP}{Center of Pressure}
\newacronym{zmp}{ZMP}{Zero Moment Point}
\newacronym{cp}{CP}{Capture Point}
\newacronym{grfs}{GRFs}{Ground Reaction Forces}
\newacronym{vhsip}{VHSIP}{Variable Height Springy Inverted Pendulum}
\newacronym{ik}{IK}{Inverse Kinematic}
\newacronym{ocp}{OCP}{Optimal Control Problem}
\newacronym{nlp}{NLP}{Nonlinear Programming}
\newacronym{ltv}{LTV}{Linear Time Varying}
\newacronym{lti}{LTI}{Linear Time Invariant}
\newacronym{td}{TD}{Touch Down}
\newacronym{lc}{LC}{Landing Controller}
\newacronym{qp}{QP}{Quadratic Program}
\newacronym{rl}{RL}{Reinforcement Learning}
\newacronym{mpc}{MPC}{Model Predictive Control}
\newacronym{wbc}{WBC}{Whole Body Control}
\newacronym{fwp}{FWP}{Feasible Wrench Polytope}
\newacronym{micp}{MICP}{Mixed-Integer Convex Program}
\newacronym{srbd}{SRBD}{Single Rigid Body Dynamics}
\newacronym{to}{TO}{Trajectory Optimization}
\newacronym{ddp}{DDP}{Differential Dynamic Program}
\newacronym{nn}{NN}{Neural Network}
\newacronym{pd}{PD}{Proportional Derivative}
\newacronym{dofs}{DoFs}{Degrees of Freedom}
\newacronym{ode}{ODE}{Ordinary Differential Equation}
\newacronym{msd}{MSD}{Mass-Spring-Damper}
\newacronym{pwbc}{pWBC}{projection-based Whole Body Control}
\newacronym{sp}{SP}{Support Polygon}
\newacronym{imu}{IMU}{Inertial Measurement Unit}
\newacronym{lip}{LIP}{Linear Inverted Pendulum}
\newacronym{haa}{haa}{hip abuction/adduction}
\newacronym{hfe}{hfe}{hip flexion/extention}
\newacronym{kfe}{kfe}{knee flexion/extention}
\newacronym{udp}{UDP}{User Datagram Protocol}
\newcommand{\MF}[1]{\textcolor{red}{\textbf{mfocchi}: #1}}

\newcommand{\FR}[1]{#1}

\newtheorem*{problem}{Problem Statement}

\newcommand{\Rnum}{\mathbb{R}} 

\DeclareMathOperator*{\argmin}{\arg\!\min}				

\makeglossaries 
\begin{document}
	\title{Reactive Landing Controller for Quadruped Robots}
	
	\author{
		Francesco Roscia$^{1}$, Michele Focchi$^{1, \, 2}$, Andrea Del Prete$^{3}$, Darwin G. Caldwell$^{4}$, and Claudio Semini$^{1}$ 
		\thanks{
			This work was supported by the co-financing of the European Union FSE-REACT-EU, PON Research and Innovation 2014-2020 DM1062/2021. The authors are with:\\
			$^1$ Dynamic Legged Systems (DLS), Istituto Italiano di Tecnologia (IIT), Genoa, Italy. \href{mailto:francesco.roscia@iit.it}{\texttt{firstname.lastname@iit.it}}\\
			$^2$ Department of Information Engineering and Computer Science (DISI), University of Trento, Trento, Italy. \href{mailto:michele.focchi@unitn.it}{\texttt{michele.focchi@unitn.it}}\\
			$^3$ Industrial Engineering Department (DII), University of Trento, Trento, Italy. \href{mailto:andrea.delprete@unitn.it}{\texttt{andrea.delprete@unitn.it}}\\
			$^4$ Advanced Robotics (ADVR), Istituto Italiano di Tecnologia (IIT), Genoa, Italy. \href{mailto:darwin.caldwell@iit.it}{\texttt{darwin.caldwell@iit.it}}}
	}
	
	
	
	
	\maketitle
	
	\begin{abstract}
		Quadruped robots are machines intended for challenging and harsh environments. Despite the progress in locomotion strategy, safely recovering from unexpected falls or planned drops is still an open problem. It is further made more difficult when high horizontal velocities are involved.
		In this work, we propose an optimization-based reactive Landing Controller that uses only proprioceptive measures for torque-controlled quadruped robots that free-fall on a flat horizontal ground, knowing neither the distance to the landing surface nor the flight time. 
		Based on an estimate of the Center of Mass horizontal velocity, the method uses the Variable Height Springy Inverted Pendulum model for continuously recomputing the feet position while the robot is falling. 
		In this way, the quadruped is ready to attain a successful landing in all directions, even in the presence of significant horizontal velocities.
		The method is demonstrated to dramatically enlarge the region of horizontal velocities that can be dealt with by a naive approach that keeps the feet still during the airborne stage. 
		To the best of our knowledge, this is the first time that a quadruped robot can successfully recover from falls with horizontal velocities up to $3 \ \mathrm{m/s}$ in simulation.
		Experiments prove that the used platform, Go1, can successfully attain a stable standing configuration from falls with various horizontal velocities and different angular perturbations. 
	\end{abstract}
	
	
	\section{Introduction}\label{sec:introduction}
	\IEEEPARstart {L}{egged} robots are designed to traverse rough terrains.
	Thanks to the progress of the last decades, they have become lighter and stronger, enabling agile locomotion.
	Many types of gaits, such as trotting or crawling, have been investigated and successfully developed for quadrupedal robots. 
	In contrast to advances in locomotion, relatively little research has been done on safely recovering after unexpected falls or planned drops. 
	These abilities can be beneficial for navigating harsh environments and preventing significant damage. 
	High horizontal velocity makes the landing problem much more challenging. 
	There are many situation where this is not negligible, e.g., when a quadruped trots almost at its maximal speed and must do a leap without first stopping the motion. 
	So, for a robot to be able to attains a stable standing posture after a fall, the control framework should estimate and deal with both the vertical and horizontal velocity components.
	Animals with righting reflexes have inspired many previous works on improving robotic landing capabilities, such as cats \cite{kane1969dynamical} and squirrels \cite{hunt2021acrobatic}.
	These works focus on dorso-ventrally reorienting the main body so that the limbs point toward the ground, e.g., mimicking cats with a flexible spine \cite{shields2013falling}.
	Some authors explored the inertial effects of flywheels, tails, and limbs for reorienting the robot when it falls. 
	The work presented in \cite{kolvenbach2019towards} used a flywheel to control the tilt angle effectively; \cite{roscia2022orientation} proved that two flywheels with intersecting rotational axes can actuate both roll and pitch, allowing for non-planar jumps and landings.
	Many researchers have introduced a tail, which is an additional link that rotates about an axis that does not pass through the robot \gls{com} \cite{wenger2016frontal, chu2019null, tang2023towards}.
	However, a tail can only perform limited corrections due to its restricted range of motion and it adds weight and complexity to the robot.
	In \cite{kurtz2022mini} Mini Cheetah was provided with heavy boots to improve the influence on torso rotation. 
	Purely vertical falls with large rotations on the sagittal plane are handled by a combination of off-line optimization and supervised machine learning. 
	The drawback of increasing the limbs' inertia is that the robot becomes tailored for the specific task of landing since planning problems can no longer rely on the assumption of mass-less legs.
	%
	\begin{figure}[t]
		\centering
		\includegraphics[width=0.83\linewidth]{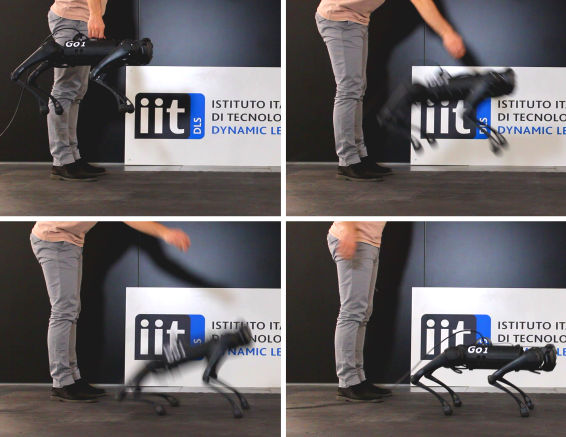}
		\caption{ Based only on proprioceptive measures, during a fall the proposed \glsfirst{lc} adjusts the limbs posture in order to drive the robot to stable standing configuration after \glsfirst{td}, avoiding bounces, feet slippage and undesirable contacts with the ground. }
		\label{fig:real_robot_snap}
		\vspace{-1.5em}
	\end{figure}
	\begin{figure*}[t]
		\centering
		\subfloat[\label{fig:lc_framework}]{
			\includegraphics[height=5.3cm]{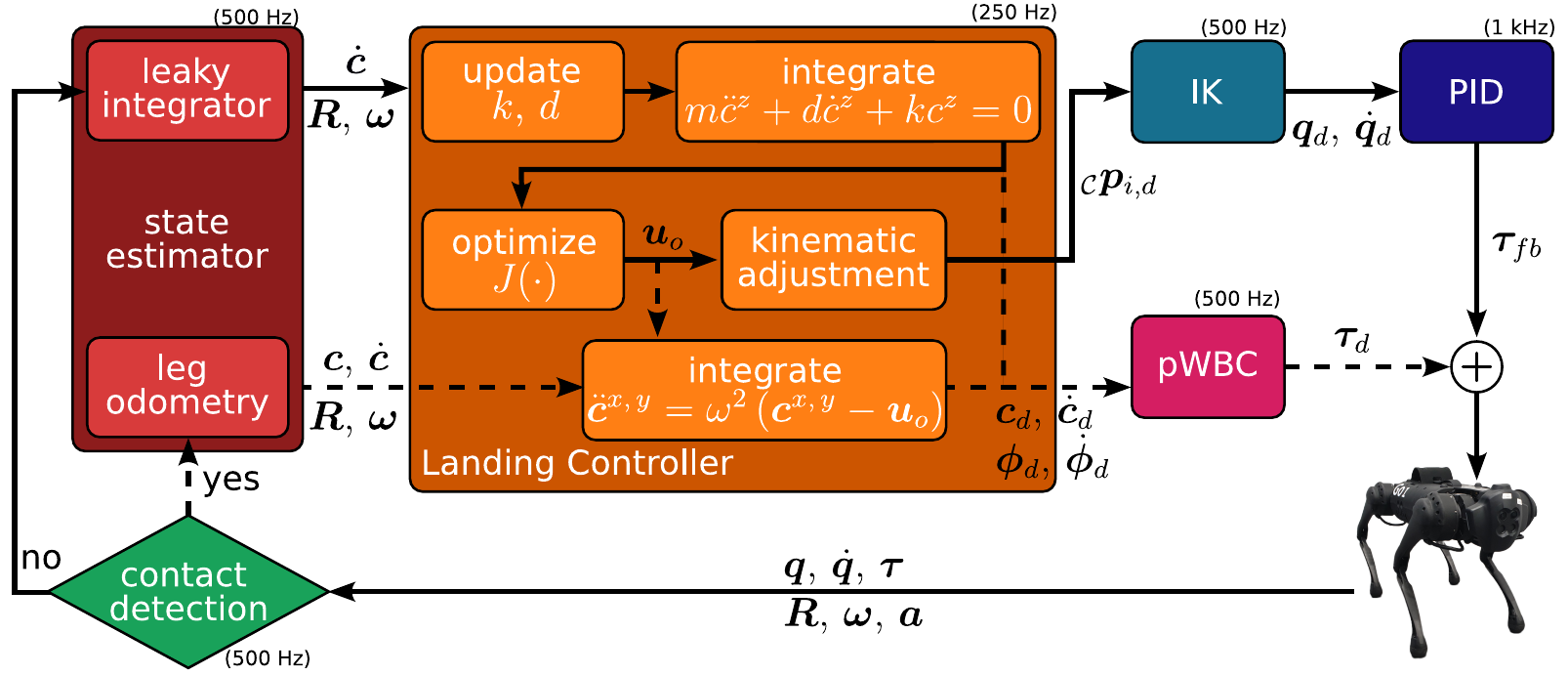}}
		\hspace{3em}
		\subfloat[\label{fig:slip3d}]{%
			\includegraphics[height=5.3cm]{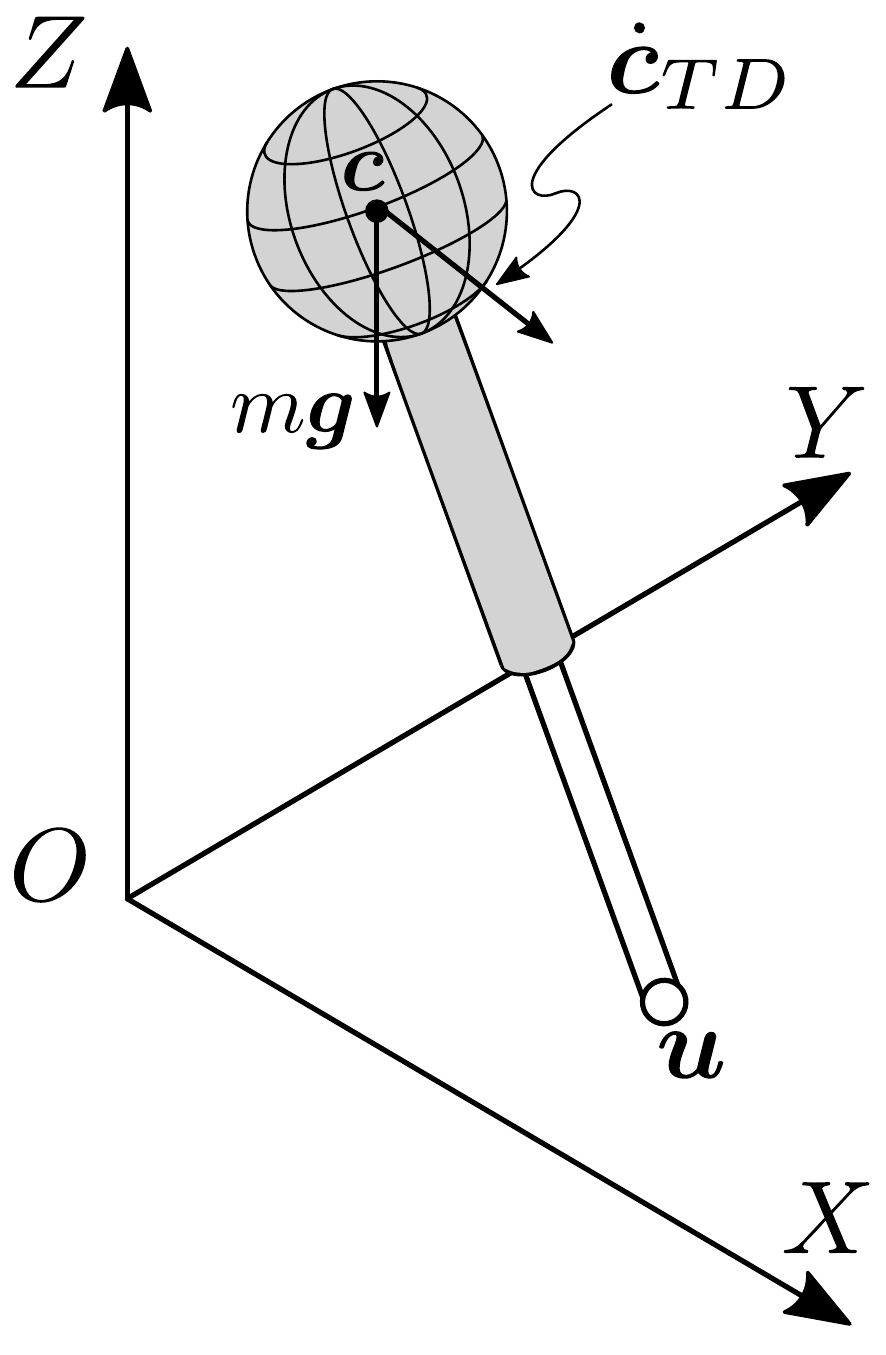}}
		\caption{(a) Overview of the \gls{lc} framework. When the robot is in the flying phase, the virtual foot location is continuously re-optimized using the template model \gls{vhsip}, and the feet are shifted accordingly. Once landing is detected, the last computed trajectory for the \gls{com} and trunk orientation is tracked. The dashed branches are active only after \gls{td} is detected. %
			(b) The 3D \gls{vhsip} model used to describe the robot dynamics after \gls{td}.}
		\label{fig1}\vspace{-1.4em}
	\end{figure*}
	%
	%
	There is copious literature on approaches based on \gls{to} to perform airborne maneuvers.
	Many of these techniques neglect the landing phase and assume perfect tracking of the optimized trajectory. 
	Minor errors in timing can complicate the execution of the landing task. 
	Using optimized joint torques as feed-forward commands and \gls{pd} joint feedback for tracking the optimized joint trajectories, flips were accomplished by Mini Cheetah in \cite{katz19mini}.
	\cite{nguyen2019optimized} and \cite{nguyen2022contact} introduced a framework for performing highly dynamic jumps by combining off-line contact timings, off-line whole-body \gls{to}, and high-frequency tracking controller. 
	Despite the relevant results shown, those approaches may struggle in different occasions. 
	E.g., if the controller does not track exactly the references before the take off, the actual aerial stage may last more/less than planned, resulting in a landing trajectory that is not feasible.
	Moreover, these methods require to compute in advance the whole references, that depend on the take off conditions. 
	For on-line optimization, such data may not be available if the robot is running and it is requested to jump without stopping.
	The MIT Mini Cheetah can perform a rich set of aerial movements, as shown in \cite{chignoli2021online}.
	Motions are planned via centroidal momentum-based nonlinear optimization and then tracked using a variational-based optimal controller \cite{chignoli2020variational}. 
	Landing is accomplished using a simple joint \gls{pd} controller stabilizing a fixed landing position. 
	Therefore, only small heights are considered. 
	Several works based on \gls{rl} and \gls{mpc} present interesting landing strategies. 
	Using a deep \gls{rl} algorithm and performing sim-to-real transfer via domain randomization, Rudin et al. show Spacebok performing consecutive 2D jumps in a micro-gravity environment \cite{rudin2021cat}. 
	Even if the method is tailored to uneven terrains, tests are conducted on flat and rigid ground. 
	A Vicon system provides measures for the robot’s absolute position and orientation.
	How to extend the results to real-world 3D conditions with more dramatic impacts and inertial effects is unclear. 
	Also \cite{qi2022integrated} addresses the problem of landing on celestial bodies. 
	It proposes a model-free \gls{rl} controller with an auto-tuning reward function to address attitude and landing control on Near-Earth asteroids for which gravity cannot be supposed to be a central force. 
	In \cite{bledt2020regularized}, MIT Mini Cheetah is dropped with a \gls{com} height of $1 \ \mathrm{m}$ above the ground. The heuristics-guided \gls{mpc} allows the robot to push onto the ground enough to arrest its downward momentum and return to a stable standing state within two steps.
	A recent work \cite{zhang23} implements heuristics to shift the feet posture for landing based on physical intuitions.
	However, the approach lacks feasibility guarantees grounded on a model of the system.
	In \cite{jeon2022online}, a complete control framework can select optimal contact locations and timings in real-time. 
	The controller consists of a \gls{nn} providing warm-start trajectories to a kino-dynamic \gls{to} problem, solved on-line in an \gls{mpc} fashion. 
	The robot safely recovers from drops 
	with different orientations, but involving only \emph{vertical} falls.
	The bottleneck of this framework is the \gls{nn}: if a solution is not found within $300 \ \mathrm{ms}$, a new request is made and the optimization restarts. 
	However, such time span represents an extensive flying phase: it corresponds to a height of about $0.5 \ \mathrm{m}$, starting with zero velocity. 
	Therefore, falls from reduced heights may not be addressed with this approach. 
	
	\subsection{Proposed Approach}
	\label{subsec:approach}
	This paper addresses the problem of landing on a flat horizontal surface for a quadruped robot. 
	Specifically, the objective is to define a control strategy that satisfies the following requirements once the robot touches the ground: 
	\begin{enumerate}[start=1,label={\bfseries ({R\arabic*})}]
		\item \label{req:bounces} no bounces after landing;  
		\item \label{req:crush} the trunk must not hit the ground;
		\item \label{req:final} the robot must reach a stable standing state;
		\item \label{req:slip} once landed, the feet must not slip.
	\end{enumerate}
	To meet these requirements, we model the dynamics of the quadruped after the touch down as a \glsfirst{vhsip}. 
	The use of a spring model is not novel in the literature \cite{raibert1986legged, blickhan1989spring, xiong2018bipedal}.
	Despite its low complexity, the template model captures the linear dynamics of the robot with sufficient accuracy and is inherently suitable for damping impacts with the terrain. 
	It allows us to design a control architecture with a short computation time, 
	keeping it adequately reactive to change the control strategy at \gls{td} promptly.
	Our approach relies only on proprioceptive measurements, so a motion capture or visual perception system is not required.
	
	\subsection{Contributions and Outline}
	The main contributions of this work are as follows.
	\begin{itemize}
		\item We introduce a real-time \emph{omni-directional} \gls{lc} framework (Fig. \ref{fig:lc_framework}), which can handle horizontal velocity, is tolerant to \gls{td} timing uncertainties, and does not need an estimate of the robot absolute position.
		\item We demonstrate successful landings in both a simulation environment and with real hardware from various heights and significant horizontal velocity, up to $3 \ \mathrm{m/s}$ dropping the robot from a height of $1 \ \mathrm{m}$. To our best knowledge, this is the first time a quadruped robot can recover from a fall with such horizontal velocity relying only on proprioceptive measures.
		\item We present a detailed analysis showing advantages and limitations of our \gls{lc}. We verify that it outperforms a naive approach that does not move the feet when the robot is falling. Although the template model neglects angular dynamics, our approach can tolerate drops starting with non-negligible orientations, e.g., with roll between $-40 ^\circ$ and $30 ^\circ$, or pitch rate from $-440 ^\circ/s$ to $210 ^\circ/s$ .
	\end{itemize}
	Even though the approach was developed and tested for four-legged robots, it is generic enough to  be easily extended to robots with any number of legs.
	%
	%
	
	The remainder of this manuscript is organized as follows. 
	In Section~\ref{sec:problemFormulation}, we discuss the \gls{vhsip} template model and formally state the landing problem. 
	The structure of the landing framework is detailed in Section~\ref{sec:methodology}, and the motion control in Section~\ref{sec:motionControl}. 
	Finally, we present implementation details and results in Section~\ref{sec:results} and we draw the conclusions in Section~\ref{sec:conclusion}.
	
	\section{Modeling and Problem Formulation}
	\label{sec:problemFormulation}
	The fall of a legged robot can be decomposed in two phases: a flying phase, in which the robot is subjected only to gravity, and a landing phase, which begins when limbs (typically the feet) come into contact with the environment and generate \acrfull{grfs}. 
	The switching instant between the two phases is named \glsfirst{td}. In this section, we derive the  \gls{vhsip} model that we use as a template to formalize and address the landing problem. 
	
	\subsection{Derivation of the \gls{vhsip} Model}
	\label{subsec:model}
	Let us introduce an inertial coordinate frame $\mathcal{W}$: the $Z$--axis is orthogonal to a flat ground and the $XY-$plane lies on it. 
	In this frame, the robot can be seen as a single rigid body of mass $m \in \mathbb{R}^+$, lumped at its \gls{com} ${\bm{c} = \begin{bmatrix} c^x & c^y & c^z \end{bmatrix}^T} \in \mathbb{R}^3$, having inertia $\tensor[_{\mathcal{C}}] {\bm{\mathcal{I}} }{_{\mathcal{C}}}  \in \mathbb{R}^{3 \times 3}$. 
	When the robot is falling, the balance of moments constrains the linear dynamics to follow the ballistic trajectory while conserving the angular momentum $\bm{L}\in\mathbb{R}^3$. 
	If the robot has $n_c$ contact points with the ground, the balance of linear and angular moments can be written as
	\begin{equation}
	m\left(\ddot{\bm{c}} + \bm{g}\right) = \sum_{i=1}^{n_c} \bm{f}_i   \qquad
	\dot{\bm{L}} = \sum_{i=1}^{n_c} \left(\bm{p}_i - \bm{c} \right) \times \bm{f}_i
	\end{equation}
	being $\bm{g} \in \mathbb{R}^3$ the (constant) gravity acceleration vector, and $\bm{p}_i\in \mathbb{R}^3$ is the position of the $i$-th contact point on which the environment exerts the force $\bm{f}_i\in \mathbb{R}^3$. 
	Under the assumptions of horizontal ground (${p_i^z = 0,} \ {\forall i=1, \, \dots, \, n_c }$) and negligible variation of the angular momentum (${\dot{\bm{L}} \approx 0}$), we obtain the relationship (for further details, see \cite{wieber2016modeling})
	\begin{equation}
	\ddot{\bm{c}}^{x, \, y} = \omega^2(t) \left(\bm{c}^{x, \, y} - \bm{u} \right),
	\label{eq:SLIP_dyn_xy}
	\end{equation}
	where $\bm{u} \in \mathbb{R}^2$ is the \gls{cop}, defined as 
	\begin{equation*}
	\bm{u} \triangleq \dfrac{\sum_{i=1}^{n_c} f_i^z \bm{p}_i^{x,y} }{\sum_{i=1}^{n_c} f_i^z}.
	\end{equation*}
	The dynamics in the horizontal directions of the \gls{com} are decoupled, but they depend on the vertical motion through 
	\begin{equation*}
	\omega^2 \triangleq \left(g^z + \ddot{c}^z\right) / c^z.
	\end{equation*}
	%
	Let us assume the \gls{cop} to be constant, as a \emph{virtual foot}.
	The vector $\overrightarrow{\bm{c} \bm{u}}$ can change its length and rotate about $\bm{u}$ due only to the gravity force $m\bm{g}$ and the initial \gls{com} velocity $\dot{\bm{c}}^{x, \, y}_{0}$ (for our case it equals the \gls{td} velocity $\dot{\bm{c}}^{x, \, y}_{TD}$, see Fig. \ref{fig:slip3d}).
	%
	%
	If the \gls{com} height is constant, \eqref{eq:SLIP_dyn_xy} simplifies to the well-known \gls{lip} model \cite{kajita1991study}. Nevertheless, such assumption is too restrictive. 
	Indeed, to achieve a smooth landing and dissipate the impact energy effectively, it is convenient to enforce the vertical \gls{com} dynamics after \gls{td} to behave as a \gls{msd} system:
	\begin{equation}
	m \ddot{c}^z + d \dot{c}^z + k (c^z-l_0) = F^z,
	\label{eq:msd} 
	\end{equation}
	in which $l_0$ is the spring rest position, and $k,\, d \in \mathbb{R}^+$ are the \emph{virtual stiffness} and \emph{virtual damping} coefficients, respectively. 
	If the joint controller already implements a gravity compensation strategy, one may assume $F^z = 0$. In this way, the vertical \gls{com} dynamics is an autonomous system, 
	depending only on the state at \gls{td}: $c^z_{TD}$ and $\dot{c}^z_{TD}$. 
	We will refer to \eqref{eq:SLIP_dyn_xy}--\eqref{eq:msd} as the \gls{vhsip} model.
	Since the CoM dynamics of the \gls{vhsip} in the two horizontal directions are equivalent and decoupled, 
	we analyze only the motion along the $X-$ axis, keeping in mind that the arguments are valid also  along the $Y-$ axis.
	
	After \gls{td} the system is associated with a conserved quantity, the so-called Orbital Energy $E(c^x-u^x, \, \dot{c}^x)$ \cite{pratt2007derivation}:
	\begin{equation*}
	E = \dfrac{1}{2} \dot{r}^{x^2} h^2(r^{x}) + g^z r^{x^2} f(r^{x}) -3g^z \int_0^{r^{x}} f(\xi)\xi d\xi
	\end{equation*}
	with $r^x = c^x-u^x$, $f$ is a twice-differentiable function
	\footnote{The scalar function $f$ maps horizontal displacements to heights. It exists as long as $c^x(t)$ is a bijective function of time. As a matter of fact, in this case $c^x(t)$ admits an inverse which would allow us to write $c^z(t)$ as a function of $c^x(t)$. The trajectory $c^x(t)$ is bijective if the virtual foot $u^x$ is constant and $\omega^2(t)>0$. The former clause is already taken as an assumption. The latter occurs whenever the \gls{com} does not penetrate the ground ($c^z>0$) and there is no pulling force from the ground ($\ddot{c}^z>-g^z$): this clause is always verified for the problem we are considering. }
	for which it holds $c^z = f(r^x)$ and $f'$ is its derivative. Additionally, we define $h(r^{x}) = f(r^{x}) - f'(r^{x})r^{x}$. 
	If the \gls{com} is moving toward the virtual foot with $E>0$, then orbital energy is sufficient to let the \gls{com} to pass over $\bm{u}$ and continue on its way. 
	If $E<0$, then the \gls{com} will stop and reverse the direction of motion before getting over the virtual foot. 
	If $E=0$, then the CoM converges to rest above $\bm{u}$, \cite{pratt2006capture}. 
	Therefore, robot configurations in equilibrium can be reached by selecting $\bm{u}$ so that at \gls{td} the following condition holds
	\begin{equation}
	\label{eq:zero-orbital-energy}
	E(c^x_{TD}-u^x, \, \dot{c}^x_{TD})=0.
	\end{equation}
	In the capturability framework, such $\bm{u}$ is named \gls{cp}.
	Whereas for the \gls{lip} model it is possible to explicitly compute it, for our \gls{vhsip} model this is not the case. 
	In the following, we show how we overcome this complexity.
	
	\subsection{Landing Problem }
	Now, we can state the problem more formally.
	\begin{problem}
		Consider a robot that is free falling on a flat horizontal ground with
		negligible angular velocity. Without knowing the distance between the robot and the ground, find the parameters for the template \gls{vhsip} model (i.e., $k$, $d$ and $\bm{u}$), that fulfill the requirements \ref{req:bounces}, \ref{req:crush},
		\ref{req:final}, and \ref{req:slip}.
		Then, compute the joint torques that realize the \gls{com} motion obtained with the selected template model.
	\end{problem}
	
	\section{Methodology}
	\label{sec:methodology}
\begin{figure}[t!]
		\centering
		\includegraphics[width=.6\linewidth]{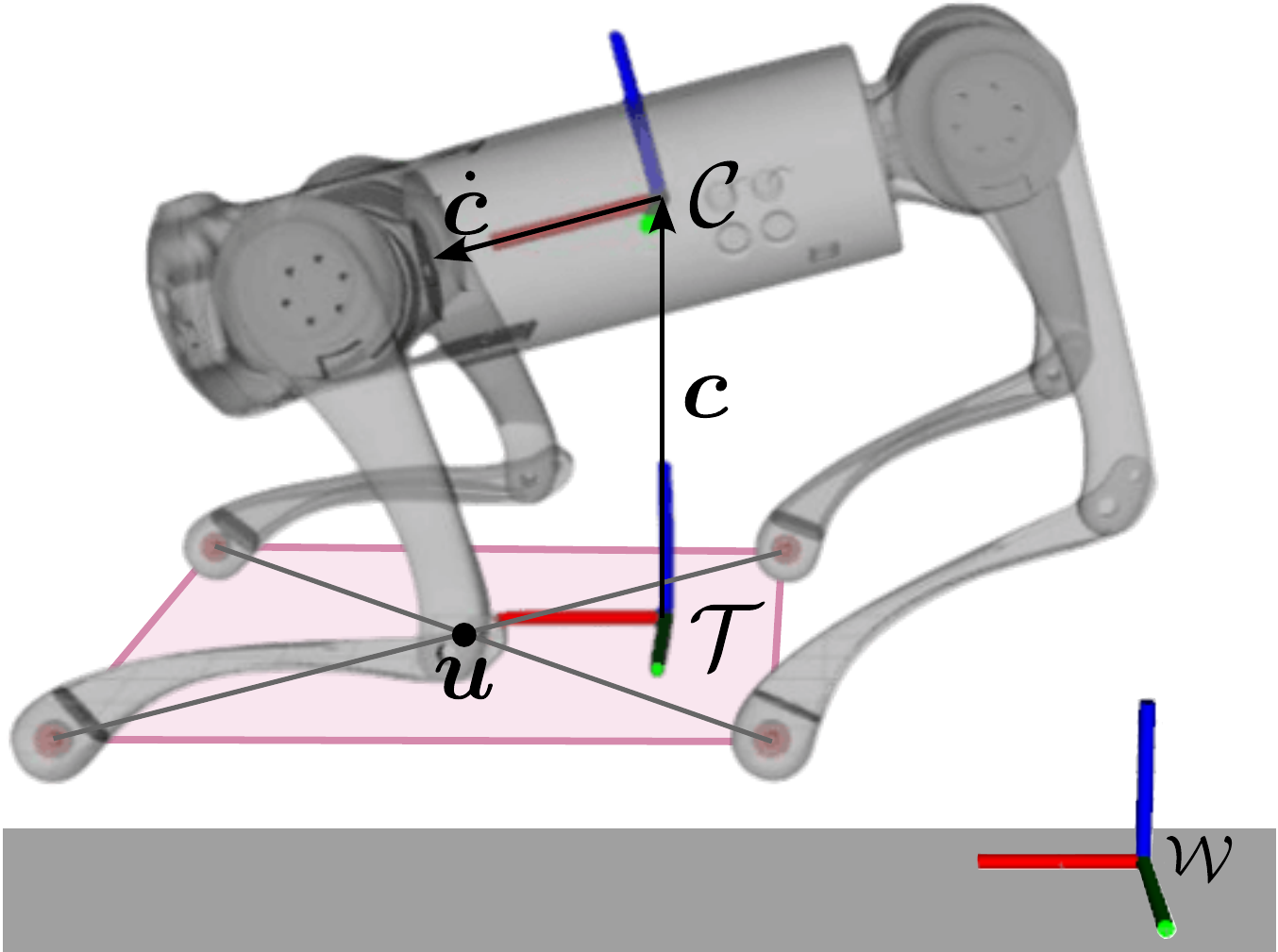}
		\caption{The \gls{com} frame $\mathcal{C}$ has the origin at the robot \gls{com} and encapsulates the trunk orientation. The terrain frame $\mathcal{T}$ is the horizontal reference frame for which the \gls{com} has coordinates $\bm{c}=\begin{bmatrix} 0 & 0 & l_0\end{bmatrix}^T$ when the robot is in air. After \gls{td}, $\mathcal{T}$ is kept fix. Respectively, red, green and blue segments denote the $X$, $Y$, and $Z$ axes.
			The robot is performing the kinematic adjustment since its orientation is not horizontal and $\dot{\bm{c}}$ has a non-null horizontal component.
		}
		\label{fig:kin_adj}
		\vspace{-1.4em} \end{figure}

	In this section, the structure of our \gls{lc} is introduced and discussed. At any sampling instant, we suppose that the \gls{td} is about to occur and use the template \gls{vhsip} model to compute a new \gls{com} reference trajectory for landing. 
	In this way, we can avoid estimating the robot's absolute position when no contact is active.
	During the flying phase, the system has to prepare to dissipate the kinetic energy throughout the landing phase.
	The robot should adjust its limbs to be able to place the virtual foot $\bm{u}$ (i.e., the \gls{cop}) on the \gls{cp} at \gls{td}. 
	Notice that \eqref{eq:SLIP_dyn_xy}--\eqref{eq:msd} are defined in an inertial frame, but our \gls{lc} is designed to be independent of absolute position estimates. 
	We circumvent this limitation by introducing the terrain frame $\mathcal{T}$.
	First, we denote the robot \gls{com} frame with $\mathcal{C}$ having the origin on the \gls{com}, the $X_\mathcal{C}$--axis along the forward direction of the main body of the robot and the $Y_\mathcal{C}$--axis orthogonal to it, pointing towards the left side (see Fig. \ref{fig:kin_adj}).
	The terrain frame is a horizontal frame \cite{barasuol2013reactive}, hence the $X_\mathcal{T}Y_\mathcal{T}$--plane is orthogonal to gravity and $X_\mathcal{T}$ and $X_\mathcal{C}$ lie on the same plane. 
	We will refer to the distance between $O_\mathcal{T}$ and $O_\mathcal{C}$ during the flying phase as $l_0$, which is a user-modifiable parameter. 
	For a given control interval, the frame $\mathcal{T}$ is fixed and can be employed to compute the new \gls{com} reference trajectory. 
	At the subsequent control interval, one of the following two alternative conditions will arise:
	\begin{itemize}
		\item \gls{td} is detected. Thus, the \gls{com} reference will be tracked, stabilizing the system.
		\item \gls{td} is not detected. A \emph{kinematic adjustment} (detailed in Section~\ref{subsec:kin_adj}) is performed, in order to keep the feet aligned with the ground, i.e., on the $X_\mathcal{T}Y_\mathcal{T}$--plane. Then, a new terrain frame is set and the process repeats.
	\end{itemize}
	The benefits of continuous re-computation are two-fold: to make the robot reactive and to avoid the need of state estimation\footnote{We still have to estimate the linear velocity, but the implementation described in Section~\ref{subsec:s_est} mitigates the influence of accelerometer biases.}.
	Since all the quantities in the remainder of this section are expressed in the terrain frame, henceforth, we will omit the frame for the sake of readability. 
	
	\subsection{Vertical Motion Reference}
	\label{subsec:VerticalMotionReference}
	In the context of the \gls{vhsip} model, the requirement \ref{req:bounces} is equivalent to having a critically-damped oscillator in \eqref{eq:msd}. 
	This can be achieved setting the damping equal to $d = 2 \sqrt{km}$. 
	Thus, the two system poles are real and coincident in ${\lambda = -\sqrt{k/m}}$. 
	The kinematic adjustment module guarantees that when the \gls{td} occurs, the \gls{com} is always at distance $l_0$ from the floor. 
	Then, the reference trajectory for the \gls{com} in the vertical direction depends only on the \gls{td} velocity  $\dot{c}^z_{TD}<0$. 
	The evolution of a critically-damped system can be computed in closed form:
	\begin{subequations}
		\begin{align}
		\label{eq:comz_v_ref}
		\dot{c}^z(t) &= e^{\lambda \left(t-t_{TD}\right)} \ \left(\lambda 
		\left(t-t_{TD}\right) + 1\right) \ \dot{c}^z_{TD}\\	
		\label{eq:comz_p_ref}
		c^z(t) &= e^{\lambda \left(t-t_{TD}\right)} \ \left(t-t_{TD}\right) \ \dot{c}^z_{TD} + l_0
		\end{align}
	\end{subequations}
	The stiffness $k$ must be selected to ensure the minimum value for $c^z(t)$ is above the ground, preventing undesired collisions between trunk and ground (requirement \ref{req:crush}).
	Having a critically-damped system, this minimum is unique for $t > t_{TD}$. 
	Therefore, denoting with $\Delta^z$ the minimum terrain clearance, the requirement $c^z(t) \geq \Delta^z $ for $t \geq t_{TD}$ can be translated in a lower-bound for the stiffness coefficient: 
	\begin{equation*}
	k  \geq k_1 =  \dfrac{m}{\big(e\left( \Delta^z - l_0 \right)\big)^2} \left( \dot{c}^z_{TD}\right)^2.
	\end{equation*}
	We set a maximum for the convergence time to prevent long dynamic evolution when \gls{td} velocity is low.
	Being \eqref{eq:msd} a stable second-order system with coincident eigenvalues, it is at steady-state for $t \geq -7/\lambda$. Choosing the maximum time of convergence $t_c$, another limitation for the stiffness comes out:
	\begin{equation*}
	k  \leq k_2 = m \left(t_c/7\right)^2.
	\end{equation*}
	At every control instant, we fix the virtual stiffness to $k=\max\{k_1, \, k_2\}$ and update the virtual damping accordingly. 
	Then, a new \gls{com} reference trajectory is computed.

	\subsection{Horizontal Motion Reference}
	\label{subsec:HorizontalMotionReference}
	To prevent the robot from tipping over during landing and to stabilize it into a standing configuration \ref{req:final}, we seek for the (constant) virtual foot location $u^x$ that drives the \gls{com} above it with zero linear velocity when the system reaches steady-state, i.e., that attains zero orbital energy \eqref{eq:zero-orbital-energy}. 
	Equation \eqref{eq:SLIP_dyn_xy} can be rewritten in state-space form:
	\begin{equation}
	\label{eq:SLIP_ss}
	\underbrace{
		\left[
		\begin{array} {c}
		\dot{c}^x(t) \\
		\ddot{c}^x(t)  
		\end{array}
		\right]}_{\dot{\bm{\mathrm{x}}}(t)}  = 
	\underbrace{
		\left[
		\begin{matrix}
		0 & 1 \\
		\omega^2(t) & 0
		\end{matrix}
		\right]}_{\bm{A}(t)}
	\underbrace{
		\left[
		\begin{array} {c}
		c^x(t) \\
		\dot{c}^x(t)	
		\end{array}
		\right]}_{\bm{\mathrm{x}}(t)}+ 
	\underbrace{
		\left[
		\begin{array} {c}
		0\\
		-\omega^2(t) 
		\end{array}
		\right]}_{\bm{B}(t)}
	u^x
	\end{equation}
	showing its nature of \gls{ltv} system, with $\omega(t)$ evolving in time. 
	Even if the dynamics is linear, its integration does not admit a closed-form solution because the matrix $\bm{A}(t)$ does not commute with its integral $\int_{t_{TD}}^t \bm{A}(\sigma) d\sigma$, \cite{chen1984linear}.
	Then, 
	we set up an \gls{ocp} to find the optimal \gls{cop} position $u^x_o$ that makes the \gls{com} converge above it with null velocity over a finite horizon. 
	Hereafter, when a time-varying variable appears with as subscript, e.g., $k$, it must be interpreted as evaluated at time $k T_s$, for instance $x_k = x(k  T_s)$, being $T_s$ the sampling time. 
	Our \gls{ocp} relies on forward Euler integration and it is formulated as
	\begin{equation}
	\label{eq:nlp}
	\begin{split}
	\min_{\bm{\mathrm{x}}_k, \ u^x} \ &   w_p \lvert \bm{C}_p \bm{\mathrm{x}}_N - u^x\rvert^2 + w_v \lvert \bm{C}_v \bm{\mathrm{x}}_N \rvert^2 + w_u \lvert u^x \rvert^2 \\
	\mathrm{s.t.} \quad & \bm{\mathrm{x}}_{k+1} = \bar{\bm{A}}_k \bm{\mathrm{x}}_k + \bar{\bm{B}}_k u^x \qquad k = 0, \, 1, \, \dots, \, N-1
	\end{split}
	\end{equation}
	having set $\bar{\bm{A}}_k = \bm{I}_{2 \times 2} + T_s \bm{A}_k$ and $\bar{\bm{B}}_k = T_s \bm{B}_k$.
	$\bm{I}_{2 \times 2}$ is the $2 \times 2$ identity matrix, and
	$\bm{C}_p = \left[ \begin{array}{cc} 1 & 0 \end{array} \right]$ and $\bm{C}_v = \left[ \begin{array}{cc} 0 & 1 \end{array} \right]$ are selection matrices that pick out position and velocity from the state $\bm{\mathrm{x}}$, respectively. 
	Moreover, $w_p, \, w_v, \, w_u \in \mathbb{R}^{+}$ are weights that penalize the final \gls{com} deviation from the virtual foot, the final \gls{com} velocity, and the virtual foot distance from $O_\mathcal{T}$. 
	The horizon $N$ should be large enough to guarantee that at time $N T_s$ the system is at steady-state. 
	We solve \eqref{eq:nlp} with a single shooting approach. For doing so, the knowledge of $\omega^2(t), \, 0 \leq t \leq NT_s,$ is needed. Thus, \eqref{eq:msd} must be forward integrated up to $N T_s$. Recursively applying \eqref{eq:SLIP_dyn_xy}, we express the state trajectory as a function of the virtual foot and the initial conditions. This results in an unconstrained \gls{qp} in the scalar optimization variable $u^x$:
	\begin{equation}
	\label{eq:nlp_complressed}
	\min_{u^x} \ J(\bm{\mathrm{x}}_0, u^x) \triangleq  \bm{\mathrm{x}}_0^T \bm{Q}^x \bm{\mathrm{x}}_0 + {u^x}^T \bm{Q}^u u^x + 2 \bm{\mathrm{x}}_0^T \bm{Q}^{xu} u^x
	\end{equation}
	$\bm{Q}^x$, $\bm{Q}^u$ and $\bm{Q}^{xu}$ are real matrices of dimensions $2 \times 2$, $1 \times 1$ and $2 \times 1$, respectively. 
	For the sake of conciseness, we do not report their expression here.
	Zeroing the partial derivative of the cost $J(\cdot)$, the optimal virtual foot location is found:
	\begin{equation*}
	u^x_{o} = 
	-\dfrac{\bm{Q}^{{xu}^{T}}}{\bm{Q}^u}\bm{\mathrm{x}}_0.
	\end{equation*}
	Notice that the minimization of $J(\cdot)$ corresponds to minimizing the Orbital Energy.
	Repeating the same argument for the lateral component of the \gls{com}, 
	the coordinates of the virtual foot in the terrain frame are obtained.
	%
	If a \gls{td} occurs, the robot must track the \gls{com} reference trajectory. 
	This is obtained by plugging $\bm{u}_{o}$ into \eqref{eq:SLIP_dyn_xy} and forward integrating along the horizon to get the horizontal components, while keeping the vertical component previously computed as described in \eqref{eq:msd}.
	Conversely, if a \gls{td} does not occur, the kinematic adjustment described in Section~\ref{subsec:kin_adj} is performed, shifting the feet according to the optimal virtual foot location.
	Additionally, the terrain frame is updated and a new reference is computed. 
	\subsection{Angular Motion Reference}
	\label{subsec:angular}
	Even though the \gls{vhsip} does not capture the angular dynamics of the robot, it is still valid within a range of orientations and angular rates. Suppose the robot lands with a non-horizontal trunk. In that case, we plan the Euler angles $\bm{\phi}_d\in\mathbb{R}^3$ to reach a horizontal configuration while the second-order system in \eqref{eq:msd} attains the steady state:
	\begin{equation}
	\bm{\phi}_d(t) = e^{\lambda \left(t-t_{TD}\right)} \ \left(t-t_{TD}\right) \ \dot{\bm{\phi}}_{TD}.
	\end{equation}
	Euler rates can be used since the robot operates far enough from singular configurations. 
	
	\subsection{Kinematic Adjustment}
	\label{subsec:kin_adj}
	During the flying phase, the robot limbs must prepare to realize the optimal \gls{cop} at \gls{td}. 
	We call such motion \emph{kinematic adjustment} \cite{barasuol2013reactive}. Fig. \ref{fig:kin_adj} shows the robot while performing the kinematic adjustment.
	It consists in placing the robot feet $\bm{p}_i$ on the vertices of a rectangle parallel to the landing surface, even if the robot trunk is not horizontal. 
	The rectangle lies on the $X_\mathcal{T}Y_\mathcal{T}-$plane and it is shifted such that its centroid is placed onto the optimal \gls{cop} $\bm{u}_{o}$:
	\begin{equation}
	\bm{p}_{i, \, d}(t) = \bm{p}_{i,\, 0} + \alpha(t) \, \bm{u}_{o},
	\end{equation}
	where $\bm{p}_{i,\, 0}$ is the position of the $i$-th foot when the robot is in the homing configuration.
	To avoid abrupt changes in the reference that would affect the orientation of the base, we perform a linear interpolation with ${\alpha:t\rightarrow\left[0, \, 1\right]}$. 
	Denoting with $\tensor[_{\mathcal{C}}]{\bm{R}}{_{\mathcal{T}}}\in SO(3)$ the rotation matrix describing the orientation of the terrain frame  with respect to the trunk, we can express feet positions in the $\mathcal{C}-$frame
	$
	\tensor[_{\mathcal{C}}]{\bm{p}}{_{i, \, d}} = \tensor[_{\mathcal{C}}]{\bm{R}}{_{\mathcal{T}}} \big( \bm{p}_{i, \, d} - 
	\begin{bmatrix}
	0& 0& l_0
	\end{bmatrix}^T    \big).
	$
	%
	Feet motions can be realized with joint-space control. 
	Desired joint configurations can be computed solving an \gls{ik} problem for each foot assuming mass-less legs: $
	\bm{q}_{i, \, d} = \mathrm{IK} ( \tensor[_{\mathcal{C}}]{\bm{p}}{_{i, \, d}})
	$, where $\bm{q}_{i, \, d}$ contains the desired angles of the joints on the $i-$th leg. The mechanical structure of our robot allows us to employ closed form \gls{ik} at the position level and $\bm{q}_{i, \, d} \in \mathbb{R}^3$.
	Desired joint velocities are computed considering the Cartesian-space error ${\tensor[_{\mathcal{C}}]{\bm{e}}{_{i}} = \tensor[_{\mathcal{C}}]{\bm{p}}{_{i, \, d}} -\tensor[_{\mathcal{C}}]{\bm{p}}{_{i}}}$:
	\begin{equation*}
	\dot{\bm{q}}_{i, \, d} = \big(\bm{J}_i \left( \bm{q} \right)+\varepsilon \bm{I}_{3 \times 3}\big)^{-1} k_v \, \tensor[_{\mathcal{C}}]{\bm{e}}{_{i}}
	\end{equation*}
	being $\bm{J}_i \left( \bm{q}_{d} \right) \in \mathbb{R}^{3 \times 3}$ the Jacobian matrix associated to the $i$-th foot, $\bm{I}_{3 \times 3}$ is the $3 \times 3$ identity matrix,  $\varepsilon \in \mathbb{R}^+$ a regularization parameter, and $k_v  \in \mathbb{R}^+$ a scaling factor.
	The joint trajectory is tracked with a joint-space \gls{pd} controller
	\begin{equation}
	\bm{\tau}_{fb} = \bm{K}_{P}^{j} \left( \bm{q}_{d} - \bm{q} \right) + \bm{K}_{D}^{j} \left( \dot{\bm{q}}_{d} - \dot{\bm{q}} \right) + \bm{\tau}_g\left(\bm{q}\right).
	\end{equation}
	with the gravity compensation $\bm{\tau}_g\left(\bm{q}\right)$ as feedforward term. We tuned off-line the diagonal matrices of positive gains $\bm{K}_{P}^{j}$ and $\bm{K}_{D}^{j}$, executing tasks with the feet not touching the ground.
	
	\subsection{Touch Down Detection}
	We define the transition between flying and landing phases, the \glsfirst{td}, as the first time instant when all feet are in contact with the ground. 
	Mechanical switches or contact sensors could be employed, but these devices are typically expensive and suffer from impacts. 
	Since torque estimates based on the motor current are available on our robot, we identify the contact status of the $i$-th foot from an estimate of the \gls{grfs} $\hat{\bm{f}}_i$ exerted on it: 
	\begin{equation}
	\label{eq:force_static}
	\hat{\bm{f}}_i = \bm{J}_i(\bm{q})^{-T} \bm{S}_i \big(\bm{C} (\bm{q} ) \, \dot{\bm{q}}  + \bm{g} (\bm{q}) - \bm{\tau}\big)
	\end{equation}
	where $\bm{q}$, $\dot{\bm{q}}$ and $\bm{\tau}$ are the measured joint positions, velocities and torques, respectively, $\bm{C} (\bm{q}) \, \dot{\bm{q}}$ is the centrifugal and Coriolis term, $\bm{g}(\bm{q})$ is the vector of gravitational effects, $\bm{S}_i$ is the matrix picking out torques associated to the $i$-th leg. 
	Because of the flat horizontal ground assumption, a contact is detected if $\hat{f}_i^z \geq f_{th}^z$, with $f_{th}^z$ a (small) robot-dependent threshold. 
	%
	%

	\subsection{State Estimation}
	\label{subsec:s_est}
	Because of the terrain frame definition and of the kinematic adjustment performed in the flying phase, we do not need to estimate the position of the \gls{com} at \gls{td}: we can assume it to be $\bm{c}_{TD} = \begin{bmatrix}
	0 & 0 & l_0
	\end{bmatrix}^T$. 
	Our robot is equipped with an \gls{imu} that outputs rotation matrix $\tensor[_{\mathcal{W}}]{\bm{R}}{_{\mathcal{C}}}$, angular velocity ${}_{\mathcal{C}}\bm{\omega}$ and linear acceleration ${}_{\mathcal{C}}\bm{a}$ (referred to the robot base frame).
	The linear acceleration in the inertial frame is retrieved considering the accelerometer bias $_\mathcal{C}\bm{b} \in\mathbb{R}^3$ as: $\bm{a} = \tensor[_{\mathcal{W}}]{\bm{R}}{_{\mathcal{C}}} \left( _\mathcal{C}\bm{a} - _\mathcal{C}\bm{b} \right) - \bm{g}$.
	During the flying phase, we reconstruct the base linear velocity $\bm{v}\in \mathbb{R}^3$ making use of a leaky integrator, equally to \cite{herzog2014balancing}:
	\begin{equation}
	\hat{\bm{v}}_{k+1} = 
	\left(\bm{I}_{3 \times 3} - \bm{\Gamma} T_s \right) \hat{\bm{v}}_{k} 
	+ \bm{I}_{3 \times 3} \, T_s \bm{a}_{k}
	\end{equation}
	where $\bm{\Gamma} = \mathrm{diag}\left( \gamma^x, \, \gamma^y, \, \gamma^z \right)$ is a diagonal matrix of  positive discount factors in the three directions.
	Since the flying phase has a short time duration, the leaky integrator achieves satisfactory performances in mitigating the drift due to inaccuracies of bias estimation.
	The estimate $\hat{\bm{v}}_k$ is plugged into the robot full model for computing the \gls{com} velocity during the flying phase.
	Conversely, after \gls{td} we rely on leg odometry to estimate the \gls{com} position and velocity, as in \cite{focchi2017high}.
	
	\subsection{Motion Control During Landing Phase}
	\label{sec:motionControl}
	In this section, we discuss the \gls{pwbc} we employ to track in the landing phase.
	First, we design a Cartesian impedance, attached  at the \gls{com}, to track the \gls{com} reference after \gls{td}, stabilize the orientation, and reject disturbances on both linear and angular directions. 
	The control law will generate a wrench $\bm{w}_{d}\in \mathbb{R}^6$ that we map into desired \gls{grfs} $\bm{f}_{i, \, d} \in \mathbb{R}^3$, at the robot's feet \cite{focchi2017high}.
	\subsubsection{Feedback Wrench}
	A PD feedback term is computed to track the \gls{com} reference and the base orientation:
	\begin{equation*}
	\begin{split}
	\bm{w}_{fb}^{lin} & = \bm{K}_{P}^{lin} \left( \bm{c}_{d} - \bm{c} \right) + \bm{K}_{D}^{lin} \left( \dot{\bm{c}}_{d} - \dot{\bm{c}} \right), \\
	\bm{w}_{fb}^{ang} & = \bm{K}_{P}^{ang} \,
	\bm{e}
	\left(\tensor[_{\mathcal{W}}]{\bm{R}}{_{\mathcal{C}, \, d}} \ \tensor[_{\mathcal{W}}]{\bm{R}}{_{\mathcal{C}}^T} \right) + \bm{K}_{D}^{ang} \left( \bm{\omega}_d - \bm{\omega} \right)
	\end{split}
	\end{equation*} 
	where $\tensor[_{\mathcal{W}}]{\bm{R}}{_{\mathcal{C}, \, d}}$ is the rotation matrix  associated to $\bm{\phi}_{d}$, \mbox{$\bm{e}:SO(3)\rightarrow \mathbb{R}^3$} maps a rotation matrix to the associated rotation vector and $\bm{\omega}$ is the actual angular velocity. 
	The desired angular velocity are obtained with $\bm{\omega}_d = \bm{T} \left(\bm{\phi}_d\right) \dot{\bm{\phi}_d}$.
	
	\subsubsection{Feed-forward Wrench}
	We annihilate the effects of the gravity force on the \gls{com} through the feed-forward component 
	\begin{equation*}
	\bm{w}_{g} = \begin{bmatrix}
	m\bm{g}^T & \bm{0}_{3\times1}
	\end{bmatrix}^T.
	\end{equation*}
	To improve the tracking performances, desired accelerations enter the controller with a feed-forward term
	\begin{equation*}
	\bm{w}_{ff}^{lin} = m\ddot{\bm{c}}_{d}, \qquad
	\bm{w}_{ff}^{ang} = \tensor[_{\mathcal{T}}]{\bm{R}}{_{\mathcal{C}}} \, \tensor[_{\mathcal{C}}] {\bm{\mathcal{I}} }{_{\mathcal{C}}} \,\tensor[_{\mathcal{C}}]{\bm{R}}{_{\mathcal{T}}} \dot{\bm{\omega}}_{d}.
	\end{equation*}
	where $\dot{\bm{\omega}}_{d}$ is deduced from the desired Euler angles and rates.
	
	\subsubsection{Whole Body Controller}
	After \gls{td}, the robot has all the feet in contact with the ground and we map the desired wrench 
	%
	%
	%
	\begin{equation*}
	\bm{w}_{d} =
	\bm{w}_{fb}+\bm{w}_{g}+\bm{w}_{ff}
	\end{equation*}
	to the stack of desired \gls{grfs} $\bm{f}_{d} \in \Rnum^{3n_c}$ by solving the \gls{qp}
	\begin{equation}
	\label{eq:force_qp}
	\begin{split}
	\bm{f}_d & = \argmin_{\bm{f}_{d}} \  \dfrac{1}{2} \lVert \bm{G}\bm{f}_{d} - \bm{w}_{d} \rVert^2 \\
	\mathrm{s.t.} & \quad \lvert f_{i, \,  d}^{x, \, y} \rvert \leq \mu f_{i, \,  d}^z, 
	\quad f_{i, \,  d}^z > 0  \quad i = 1, \, \dots, \, n_c
	\end{split}
	\end{equation}
	that enforces \ref{req:slip} by means of the linearized friction cone constraints. We set the friction coefficient $\mu\in\mathbb{R}^+$ to be equal for all the contacts. $\bm{G} \in \mathbb{R}^{6 \times 3n_c}$ denotes the grasp matrix
	\begin{equation*}
	\bm{G} = 
	\begin{bmatrix}
	\cdots &
	\begin{array}{c}
	\bm{I}_{3 \times 3}\\
	\left[ \bm{p}_i - \bm{c} \right]_{\times}
	\end{array}
	& \cdots
	\end{bmatrix},
	\end{equation*}
	where $\left[ \,  \cdot \, \right]_{\times}$ is the skew-symmetric operator associated to the vector product.
	Finally, the desired torque to be exerted by the joints of the $i$-th leg is $\bm{\tau}_{i, \, d} = 
	\bm{S}_i \, \bm{C} (\bm{q}) \, \dot{\bm{q}}
	- \bm{J}_i \left( \bm{q} \right) \bm{f}_{i, \, d}$.
	\section{Results}
	\label{sec:results}
	\begin{figure*}[t] 
		\centering
		\adjustbox{valign=c}{
			\subfloat[\label{fig:plot_torta}]{
				\includegraphics[height=4.5cm]{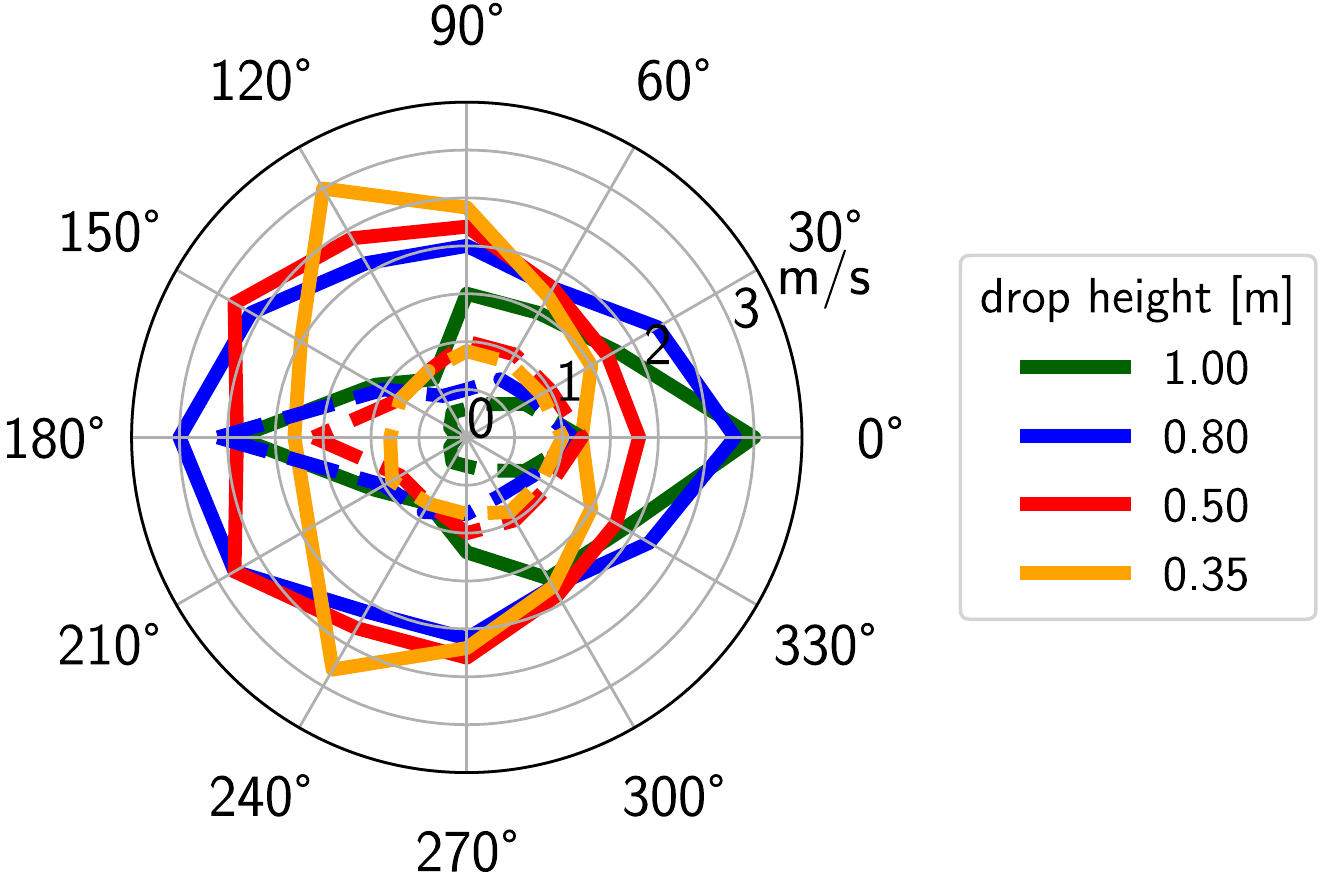}}
			\subfloat[\label{fig:colour_map}]{%
				\includegraphics[height=4.5cm]{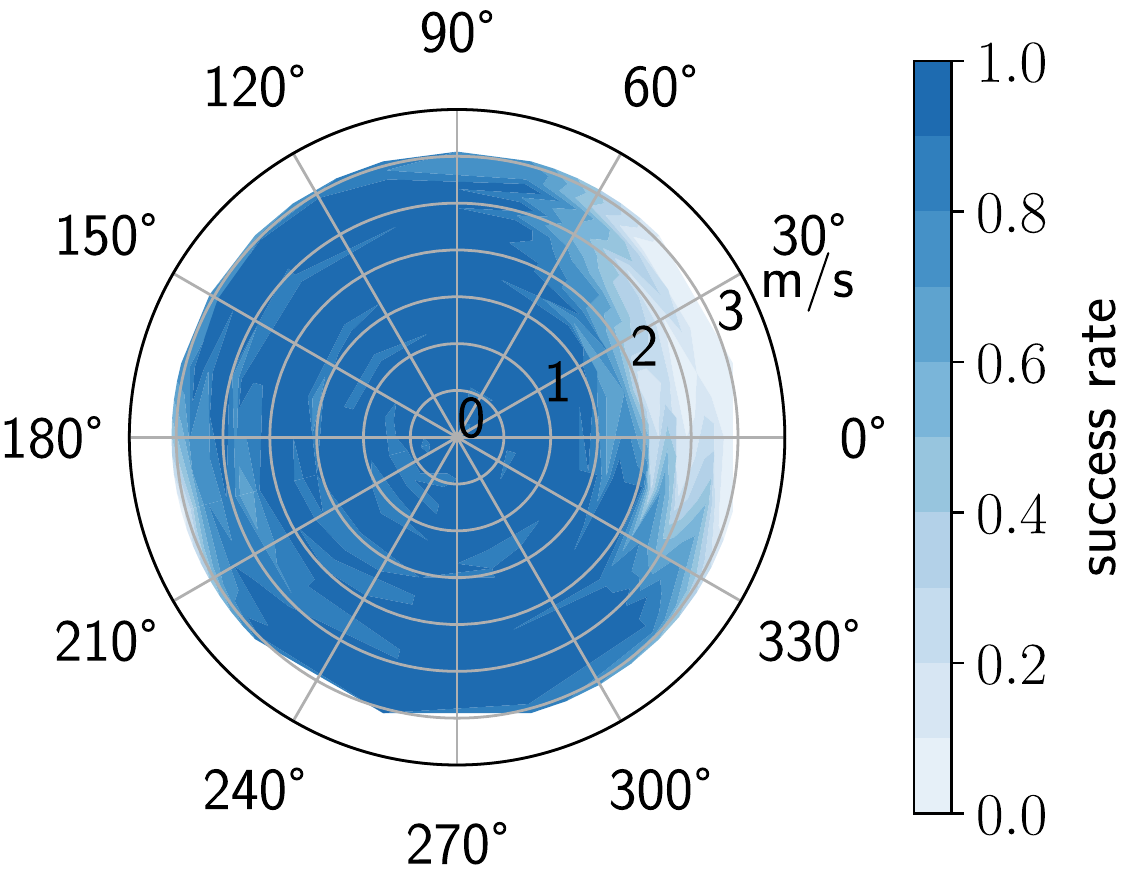}}}
		\adjustbox{valign=c}{
			{ \scriptsize
				\subfloat[\label{tab:angular_limits}]{%
					\begin{tabular}{lrrr}
						&min & max & unit \\
						\hline
						\hline
						Roll angle   &   -40 & 35 &[$^\circ$]\\
						Pitch angle &  -45 & 15 &[$^\circ$] \\
						Roll rate  &  -460 & 455 &[$^\circ/\mathrm{s}$]\\
						Pitch rate &  -365 & 215&[$^\circ/\mathrm{s}$] \\
						Yaw rate  & -3035 & 3000 &[$^\circ/\mathrm{s}$]\\
		\end{tabular} }}}
		\caption{Simulation results. (a) Polar plot of the limit magnitudes of $\dot{\bm{c}}^{x, \, y}_{0}$ for our \gls{lc} (solid lines) and for the naive approach (dashed lines) with different drop heights. The angles represent the direction of $\dot{\bm{c}}^{x, \, y}_{0}$ with respect to the $X_\mathcal{T}$-axis, i.e., $\mathrm{atan2}\left( \dot{c}^{y}_{0}, \, \dot{c}^{x}_{0}\right)$.
			(b) 
			Success rates dropping the Go1 robot from $0.80 \ \mathrm{m}$ with various initial horizontal velocities. White noise affects measured joint velocities, measured joint torques and initial horizontal velocities. 
			(c) Angular perturbations limits dropping Go1 from $0.6 \ \mathrm{m}$ height and with $1.0 \ \mathrm{m/s}$ forward velocity. The asymmetry of the robot inertia causes the roll angle and rate bounds to be not symmetric.}
		\label{fig11111}\vspace{-1.4em}
	\end{figure*}

	\subsection{Implementation details}
	The robot we use to demonstrate the validity of our approach is the torque-controlled quadruped Unitree Go1 Edu \cite{go1}.
	To visualize, simulate and interact with the robot, we use Locosim \cite{focchi2023locosim}, a platform-independent software framework designed for fast code prototyping
	\footnote{The code to replicate the results is open source and can be downloaded at 
		\href{https://github.com/iit-DLSLab/reactive\_landing\_controller/}{github.com/iit-DLSLab/reactive\_landing\_controller/}.}.
	We implemented both the \gls{pwbc} and the \gls{lc} in a Python ROS node, which relies on Pinocchio \cite{carpentier2019pinocchio} for the computation of robot kinematics and dynamics and closes the loop at \mbox{500  Hz}.
	The firmware of Go1 supports external inputs via \gls{udp} through an Ethernet connection. 
	To increase computational efficiency, we coded the \gls{vhsip} dynamics in C++ and provided Python bindings. 
	References are computed with a frequency of $250 \ \mathrm{Hz}$ and linearly interpolated to match the controller frequency of $500 \ \mathrm{Hz}$. 
	Since Python is not real-time compliant, monitoring the loop frequency in simulation is essential before running the framework with the real hardware.
	
	A number of tests are reported in the accompanying video available at \href{https://youtu.be/wiuedeHfSEY}{youtu.be/wiuedeHfSEY}.
	For all the cases, we set the parameters of Sec. \ref{subsec:VerticalMotionReference}  as follow: the nominal height $l_0 = 0.27 \ \mathrm{m}$ (robot height in home configuration),  the clearance $\Delta^z = 0.10 \ \mathrm{m}$, the maximum settling time $t_c = 1.2 \ \mathrm{s}$.

	\subsection{Limits on the Horizontal Velocity}
	To emphasize the advantages of the proposed \gls{lc}, we first compared it in simulation with a \emph{naive approach} that keeps the feet at a constant position on the $X_\mathcal{T}Y_\mathcal{T}$--plane during the flying phase. 
	The two controllers have the same reference for the \gls{com} in the vertical direction, the same joint \gls{pd} gains and the same \gls{pwbc} gains. 
	The main difference relies on the absence, in the naive approach, of the correction due to the horizontal component of the \gls{com} velocity, that it is present in our \gls{lc}.
	Here, we consider the landing task \emph{achieved} only if the feet make contact with the ground and the robot reaches a standing still configuration, i.e., joint velocities below a threshold, without bouncing.
	Our goal is to show that the modulation of the virtual foot $\bm{u}_{o}$ is crucial to achieve a successful landing. 
	By executing thousands of automated drops with varying the initial conditions of \eqref{eq:SLIP_dyn_xy}-\eqref{eq:msd}, we detected the ranges of \gls{td} horizontal velocities that can be handled using either our \gls{lc} or the naive controller, that are reported in the polar chart in Fig. \ref{fig:plot_torta}.
	We tested dropping horizontal velocities in the form $\dot{\bm{c}}^{x, \, y}_{0} = \nu \left[ \begin{array}{cc}\cos{\psi} & \sin{\psi}\end{array} \right]^T$, with $\nu = 0.0, \, 0.1 , \, \dots, 4.0 \, \mathrm{m/s}$ and $\psi = 0, \, \pi/6 , \, \dots, 2\pi  \, \mathrm{rad}$.
	For all the tested dropping heights, the \gls{lc} can handle larger initial horizontal velocities in all the directions, up to $3.0 \ \mathrm{m/s}$. 
	Notice that the maximum trotting speed of Go1 is $3.7 \ \mathrm{m/s}$, confirming our interest for landings with high horizontal velocity.
	The asymmetry of the polytopic regions results from the asymmetric mass distribution and joint limits. 
	As the naive approach often fails in simulations, we decided not to test it with any experiment because it would weaken the robot's structure.

	\subsection{Robustness to Noise}
	When dealing with real hardware it is common to have noisy measures and uncertain parameters. 
	To understand the robustness of our controller in simulation, we added white noise to measured joint velocities, measured actuation torques and initial horizontal velocity. 
	The values for the standard deviation are $\sigma_{\dot{q}} = 0.05 \ \mathrm{rad/s}$, $\sigma_{\tau} = 0.2 \ \mathrm{Nm}$ and  $\sigma_{\dot{c}^{x, \, y}} = 0.2 \ \mathrm{m/s}$, respectively.
	We throw the robot from a height of  $0.80 \ \mathrm{m}$. 
	In order to produce a statistical analysis, we execute a batch of 10 simulations for each true value of the initial horizontal velocity.
	Within each batch, we compute the success rate as the percentage of achieved landing tasks, see the colour map in Fig. \ref{fig:colour_map}.
	Despite the noise, the overall success rate is $0.875$.
	%

	\subsection{Angular perturbations}
	Here, we want to evaluate how tolerant our method is to angular perturbations, despite the fact that our template model does not describe the angular motion (see Section \ref{subsec:angular}). 
	In simulation, we tested the robustness of our \gls{lc} with respect to non-zero initial angular velocity and non-horizontal orientation of the trunk. 
	We drop the robot from a height of $0.60 \ \mathrm{m}$ (about $2.5$ times its standing height) with a forward velocity of $1.0 \ \mathrm{m/s}$. 
	One at a time, we vary the initial values of angles and rates. 
	The discretization step is $5^\circ$ for the angles and $5^\circ/\mathrm{s}$ for the rates.
	The limit values for which our \gls{lc} is able to achieve the landing task are listed in \mbox{Table \ref{tab:angular_limits}.}  
	\subsection{Experiments}
	\label{subsec:exp}
\begin{figure}[t]
		\centering
		\includegraphics[width=0.5\textwidth]{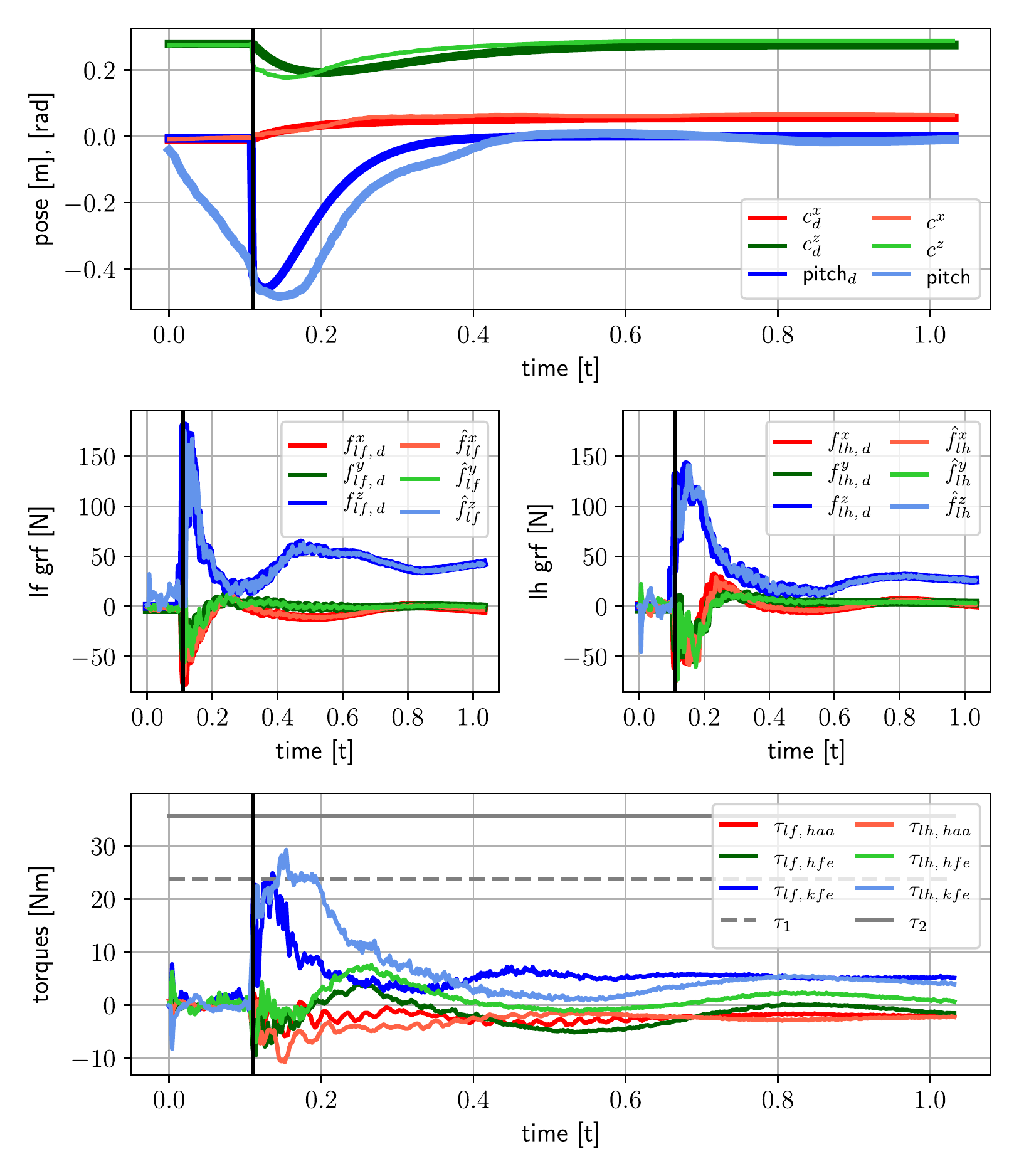}
		\vspace{-2.5em}
		\caption{Experimental results. Desired and actual \gls{com} and pitch trajectories expressed in $\mathcal{T}$-frame (top); desired and estimated \gls{grfs} exerted on the left front foot (left center) and on the left hind foot (right center); measured torques for the left legs (bottom).
			The upper actuation limit for both hip abduction/adduction (haa) and hip flexion/extension (hfe) mechanisms is $\tau_1$, for knee flexion/extension (kfe) one is $\tau_2$. 
			Lower limits are mirrored.
			The vertical lines denote the detection of the \gls{td}. }
		\label{fig:real_robot_plots}
		\vspace{-1.5em}
	\end{figure}

	%
	We performed an extensive experimental study to assess the performance of our LC. All the tests are visible in the accompanying video. We dropped the $\mathrm{12 \ kg}$ Go1 robot from various heights with different manually induced horizontal velocities in all the directions (forward, backwards, left, right). 
	Fig. \ref{fig:real_robot_snap} shows a drop from about $0.8 \ \mathrm{m}$ with non-zero horizontal velocity in the forward direction. 
	Plots in Fig. \ref{fig:real_robot_plots} are associated to this experiment.
	They illustrate that, using only proprioceptive measures, the robot can be successfully stabilized to a standing configuration thanks to both the kinematic adjustment and the \gls{grfs} exerted after landing. 
	Notice that in all the tests joint velocities and torques never reach their limits. 
	\section{Conclusions}
	\label{sec:conclusion}
	In this work, we presented a model-based approach for quadruped robot landing after unexpected falls or planned drops.
	A successful landing entails dissipating all the kinetic energy, stopping the robot without hitting the ground with the trunk and keeping the feet in contact.
	Tracking an impedance model turned out to be a suitable candidate to dissipate the excess of kinetic energy avoiding rebounds. 
	The approach is reactive enough ($500 \ \mathrm{Hz}$) to cope with heights lower than $0.5 \ \mathrm{m}$ as in \cite{jeon2022online} and it
	makes use only of proprioceptive measurements, being therefore independent of an external motion capture system. 
	Furthermore, it can achieve \emph{omni-directional} landing with significant horizontal velocity, up to $3.0 \ \mathrm{m/s}$.
	Our landing controller was extensively bench-marked in simulation and demonstrated to dramatically outperform a naive landing strategy that tracks the vertical impedance but does not adjust the feet locations during the fall.
	Despite employing a simplified model that assumes horizontal orientation, the approach was demonstrated to tolerate a large range of orientation errors.  
	An extensive experimental evaluation of omni-directional landing on the real robot Go1 was also presented, randomly dropping the robot in different ways. 
	
	Future research directions could increase the model descriptiveness, relaxing the assumption of negligible variation of the angular momentum, extending the region of feasible horizontal velocities and tolerable angular perturbations. 
	We also want to introduce a backup plan for situations in which a large horizontal velocity causes the optimal CoP to lie outside of either the kinematic region or the convex hull of the feet.
	Finally, we want to extend the approach for landing onto non-horizontal surfaces and soft terrains.

	\bibliographystyle{IEEEtran}
	\bibliography{references/root}
\end{document}